# A Decision Calculus for Belief Functions in Valuation-Based Systems

Hong XU
IRIDIA, Université libre de Bruxelles
50, Ave. F. D. Roosevelt, CP 194/6
B-1050, Brussels, Belgium
E-mail: hongxu@is1.ulb.ac.be

## Abstract

Valuation-based system (VBS) provides a general framework for representing knowledge and drawing inferences under uncertainty. Recent studies have shown that the semantics of VBS can represent and solve Bayesian decision problems (Shenoy, 1991a). The purpose of this paper is to propose a decision calculus for Dempster-Shafer (D-S) theory in the framework of VBS. The proposed calculus uses a weighting factor whose role is similar to the probabilistic interpretation of an assumption that disambiguates decision problems represented with belief functions (Strat 1990). It will be shown that with the presented calculus, if the decision problems are represented in the valuation network properly, we can solve the problems by using fusion algorithm (Shenoy 1991a). It will also be shown the presented decision calculus can be reduced to the calculus for Bayesian probability theory when probabilities, instead of belief functions, are given.

## 0. INTRODUCTION

Decision making under uncertainty is a common problem in the real world. Decision analysis provides a method for decision making. The main objective of this method is to help the decision maker to select an appropriate decision alternative in the face of uncertain environment. The traditional Bayesian decision analysis is based on the Bayesian probability theory and utility theory, where uncertain states of nature are represented by probabilities. The methods for using other uncertain calculi for decision analysis, e.g., (Jaffray 1988; Smets 1990; Strat 1990; Yager 1989) using belief functions and (Dubois & Prade 1987) using possibilities to represent the decision problems, have also been studied. Some popular methods for representing and solving decision problems are decision trees and influence diagrams.

Recently, a new framework for uncertainty representing and reasoning, called the Valuation-based System (VBS), has been proposed (Shenoy 1989,1991b). In the framework of VBS, knowledge is represented by the objects consisting of a set of variables and a set of valuations defined on the subset of variables, and the influence is drawn by using two operators called combination and marginalization. By using these objects and operators, VBS can represent uncertain knowledge in different domains including Bayesian probability theory, D-S's theory of belief functions, Zadeh-Dubois-Prade's possibility theory, etc.. More recent studies show that the semantics of VBS is also sufficient for representing and solving Bayesian decision problems (Shenoy 1991a,d). The graphical representation is called the valuation network, and the method for solving problems is called the fusion algorithm. We call the framework of VBS for Bayesian decision problem *the extended framework of VBS*. Shenoy (1991d) has also shown that the solution method of VBS is more efficient than that of decision trees and of influence diagrams.

D-S theory (Shafer 1976, Smets 1988) aims to model a decision maker's subjective valuation of evidence. Some methods have been suggested for decision analysis using belief functions. In this paper, we will develop a decision calculus for the belief functions in the extended framework of VBS. To this end, we will use a weighting factor in a way similar to the probabilistic interpretation of an assumption that disambiguates decision problems represented with belief functions (Strat 1990). We will show that with the proposed calculus, the fusion algorithm can be used for solving VBS when the problems are represented properly. We will also show that the proposed decision calculus is a kind of generalization of the calculus for Bayesian Decision problems.

The remainder of the paper is as follows: In section 1, we will describe the extended framework of VBS abstracted from the framework of VBS for Bayesian decision problems. In section 2, we will first present the decision calculus for belief functions, then give an example to show how the decision problems can be represented in the valuation network using belief functions and how the calculus benefits the fusion algorithm for solving problems through an example. Finally, in section 3, we will present our conclusions.

## 1. VALUATION-BASED SYSTEM FOR DECISION PROBLEMS

VBS (Shenoy 1991b) is a general framework for uncertainty representation and reasoning. Recent studies show that VBS can also be used for representing and solving Bayesian decision problems (Shenoy 1991a;d). In this section, we will describe the extended framework abstracted from the framework of VBS for Bayesian decision problems. In the following description, we still use the same terminology as that in VBS for Bayesian decision problems, but potentials, combination and marginalization are defined in an abstract way instead of in Bayesian probability theory.

### 1.1 VBS REPRESENTATION

A VBS representation for a decision problem is denoted by a 6-tuple $\Delta = \{\mathcal{X}_D, \mathcal{X}_R, \{\mathcal{W}_X\}_{X\in\mathcal{X}}, \{\pi_1,...,\pi_m\}, \{\rho_1,...,\rho_n\}, \rightarrow\}$, representing decision variables, random variables, frames, utility valuations, potentials, and precedence constraints, respectively. The VBS representation of a canonical decision problem $\Delta_C = \{\{D\}, \{R\}, \{\mathcal{W}_D, \mathcal{W}_R\}, \{\pi\}, \{\rho\}, \rightarrow\}$ is illustrated in Fig. 1. A graphic description is called a *valuation network*.

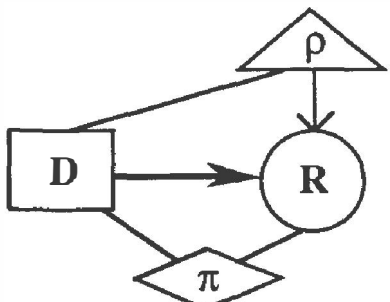

Fig 1: Graphical representation of VBS

***Variables, Frames and Configurations*** Consider a finite set of variables $\mathcal{X}$ consisting of decision variables $\mathcal{X}_D$ and random variables $\mathcal{X}_R$. The set of possible values of a variable X, denoted by $\mathcal{W}_X$, is called the *frame for X*. Given a finite non-empty subset h of $\mathcal{X}$, the *frame for h*, denoted by $\mathcal{W}_h$, is the Cartesian product of the frames for the variables in h and its elements are the *configurations of h*.

<u>Notations</u>: Let x, d, r denote the configurations, and a, b, c denote the subsets of the frames. If x is a configuration of g, y is a configuration for h, and g∩h=∅, then (x, y) denotes a configuration for g∪h. For the empty set ∅, let ♦ denote its only configuration. Thus, if x is a configuration of g, then (x, ♦)=x.

***Valuations*** Suppose $h \subseteq \mathcal{X}$. A utility (or payoff) valuation $\pi$ for h is a function from $\mathcal{W}_h$ to the set of real number. The values of utility valuations are utilities. If $X \in h$, then we say $\pi$ bears on X. Suppose $h \subseteq \mathcal{X}$ and $h \cap \mathcal{X}_R \neq \emptyset$. A potential $\rho$ for h is a function from $\mathcal{W}_h$ to [0, 1], this valuation is probability for Bayesian decision problems. In $\Delta_C$, $\pi$ is a utility valuation for {D,R}. $\rho$ will be defined later.

***Precedence constraints*** Another ingredient of the VBS representation in decision analysis is the chronology or information constraints, denoted by →. In $\Delta_C$, → is defined by D→R. Generally, four constraints are needed for the precedence relation:

(p1). The transitive closure of →, denoted by >, is a partial order (irreflexive & transitive) on $\mathcal{X}$;

(p2). For any $D \in \mathcal{X}_D$ and any $R \in \mathcal{X}_R$, either R>D or D>R;

(p3). If there is a conditional potential (explained shortly) for R ($R \in \mathcal{X}_R$) given h-{R}, and there is a decision variable $D \in h$, then D>R;

(p4). If there is a potential for h and a decision variable $D \in h$, then D>R for some random variable $R \in h$.

Apart from the ingredients of the representation, VBS have two operators called combination and marginalization for solving the problem.

***Combination*** Combination is a mapping $\otimes: \upsilon \times \upsilon \rightarrow \upsilon$, such that, if $\upsilon_g$ and $\upsilon_h$ are valuations on g and h, then $\upsilon_g \otimes \upsilon_h$ is a valuation for g∪h.

***Marginalization*** For each $h \subseteq \mathcal{X}$, there is a mapping $\downarrow h: \cup \{\upsilon_g \mid h \subseteq g\} \rightarrow \upsilon_h$, called *marginalization to h*, such that, if $\upsilon$ is a valuation for g and $h \subseteq g$, then $\upsilon^{\downarrow h}$ is a valuation on h.

With the definitions of combination and marginalization, we can now define the concepts of vacuous potential and conditional potential.

***Vacuous Potential*** Suppose $\rho$ is a potential for g. $\rho$ is *vacuous* if and only if $\upsilon \otimes \rho = \upsilon$ for all potentials $\upsilon$ for g.

***Conditional Potential*** Suppose $h \subseteq \mathcal{X}$, $R \in h$, $R \in \mathcal{X}_R$, and $\rho$ is a potential for h. $\rho$ is called a *conditional potential for R given h-{R}* if $\rho^{\downarrow h-\{R\}}$ is a vacuous potential. In $\Delta_c$, $\rho$ is a conditional potential for R given {D}.

**Strategy** The main objective in solving a decision problem is computing an optimal strategy. A strategy is a choice of an act for each decision variable D as a function of configurations of random variables R such that R>D.

**Solution for a variable** Each time we eliminate a decision variable from a valuation using maximization, we store a table of optimal values of the decision variable where the maximums are achieved. We can regard this table as a function, and call it *solution for that decision variable*. Formally, suppose $D\subseteq h$, $v$ is a valuation for h, we use $\Psi_D: \mathcal{W}_{h-\{D\}} \to \mathcal{W}_D$ to denote solution for D.

### 1.2 SOLVING VBS

In order to use VBS solution method, the VBS representation of a decision problem needs to be well-defined. Formally, the VBS representation $\Delta = \{\mathcal{X}_D, \mathcal{X}_R, \{\mathcal{W}_X\}_{X \in \mathcal{X}}, \{\pi_1, ..., \pi_m\}, \{\rho_1, ..., \rho_n\}, \to\}$ is *well-defined* if
  a. $\cup \mathcal{H}_D \supseteq \mathcal{X}_D$, where $\mathcal{H}_D$ denotes the set of subsets of $\mathcal{X}$ for which payoff valuations exist in the VBS;
  b. $\cup \mathcal{H}_R \supseteq \mathcal{X}_R$, where $\mathcal{H}_R$ denotes the set of subsets of $\mathcal{X}$ for which potentials exist;
  c. Four constraints for the precedence relation "$\to$" are satisfied;
  d. $(\rho_1 \otimes ... \otimes \rho_n)^{\downarrow q}$ is the vacuous potential for q, where q is the subset of decision variables included in the domain of the joint potential $\rho_1 \otimes ... \otimes \rho_n$.

If the VBS representation of a decision problem $\Delta$ is well-defined, then it can be reduced to an equivalent canonical problem $\Delta_c$ (Shenoy 1991d).

The main objective in solving a decision problem is computing an optimal strategy. Given a well-defined VBS representation $\Delta = \{\mathcal{X}_D, \mathcal{X}_R, \{\mathcal{W}_X\}_{X \in \mathcal{X}}, \{\pi_1, ..., \pi_m\}, \{\rho_1, ..., \rho_n\}, \to\}$, this can be achieved by two steps:

First, we compute the maximum expected value of the utilities, obtained by $(\otimes\{\pi_1, ..., \pi_m, \rho_1, ..., \rho_n\})^{\downarrow\varnothing}(\bullet)$;

Second, we compute an optimal strategy $\sigma^*$ that gives us the maximum expected value. A strategy $\sigma^*$ of $\Delta$ is optimal if $(\pi \otimes \rho)^{\downarrow\{D\}}(d_{\sigma^*}) = (\otimes\{\pi_1, ..., \pi_m, \rho_1, ..., \rho_n\})^{\downarrow\varnothing}(\bullet)$ where $\pi$, $\rho$ and D refer to the equivalent canonical problem $\Delta_c$ of $\Delta$.

Before describing the solution method, we first introduce the fusion operation: Consider a set of valuations $\alpha_1, ..., \alpha_k$, suppose $\alpha_i$ is a valuation for $h_i$. Let $\text{Fus}_X\{\alpha_1, ..., \alpha_k\}$ denote the collection of valuations after deleting a variable X, then
$$\text{Fus}_X\{\alpha_1, ..., \alpha_k\} = \{\alpha^{\downarrow(h-\{X\})}\} \cup \{\alpha_i | X \notin h_i\}$$
where $\alpha = \otimes\{\alpha_i | X \in h_i\}$, and $h = \cup\{h_i | X \in h_i\}$.

The solution method for decision problems is called the *fusion algorithm* (Shenoy 1991a). The basic idea is to successively delete all variables from the VBS. The sequence in which the variables are deleted must respect the precedence constraints, i.e. If X>Y, then Y must be deleted before X. The following theorem describes the fusion algorithm for solving VBS:

**Theorem 1:** (Shenoy 1991a) Suppose $X_1 X_2 ... X_k$ is a sequence of variables in $\mathcal{X}$ such that with respect to the partial order >, $X_1$ is a minimal element of $\mathcal{X}$, $X_2$ is a minimal element of $\mathcal{X} - \{X_1\}$, etc. Then
$$(\otimes\{\pi_1, ..., \pi_m, \rho_1, ..., \rho_n\})^{\downarrow\varnothing}(\bullet)$$
$$= \text{Fus}_{X_k}\{...\text{Fus}_{X_2}\{\text{Fus}_{X_1}\{\pi_1, ..., \pi_m, \rho_1, ..., \rho_n\}\}\}.$$

## 2. DECISION ANALYSIS USING BELIEF FUNCTIONS IN VBS

In this section, we first concrete the abstract concept of potential in the case of belief functions, then we present the operators for solution method in this case. Finally, we give an example of decision problems to show how belief functions may be used for decision analysis.

In the VBS framework for Bayesian decision problems, Shenoy (1991d) introduced a division operator involved in the solution method, he also gave some restriction for VBS representation in order to avoid divisions during computation. The decision calculus to be discussed is only for such cases when divisions can be avoided. Thus, we need the following assumption for the VBS representation.

**Assumption 1:** In the VBS representation, we have only a conditional potential for each random variable such that the variables on which the potentials are conditioned always precede the random variable.

### 2.1 POTENTIALS

In the case of D-S theory, potentials are defined as belief functions or basic probability assignments.

**Basic Probability Assignment** A *basic probability assignment (bpa)* m for h, is a function which assigns a value in [0, 1] to every subset $a$ of $\mathcal{W}_h$ and satisfies the following axioms:
(i)  $m(\varnothing) = 0$; and
(ii) $\sum\{m(a) | a \subseteq \mathcal{W}_h\} = 1$

**Belief Function** The *belief function* $\text{Bel}_m$ associated with the bpa m, is defined by:
$$\text{Bel}_m(a) = \sum\{m(b) | b \subseteq a\}$$

A subset $a$ for $m(a) > 0$ is called a *focal element* of $Bel_m$. The belief function defined by $m(W_h) = 1$ is called the *vacuous belief function* on h.

We first review the concepts of projection, extension and ballooning extension before defining combination and marginalization.

*Projection and Extension* Projection of *configurations* simply means dropping the extra coordinates. If g and h are sets of variables, $h \subseteq g$, and x is a configuration of g, then let $x^{\downarrow h}$ denote the projection of x to $W_h$. $x^{\downarrow h}$ is a configuration of h. If $h = \emptyset$, then $x^{\downarrow h} = \bullet$. If $g$ is a non-empty subset of $W_g$, then the *projection of $g$ to h*, denoted by $g^{\downarrow h}$, is obtained by $g^{\downarrow h} = \{x^{\downarrow h} \mid x \in g\}$. If $h$ is a subset of $W_h$, then the *extension of h to g*, denoted by $h^{\uparrow g}$, is $h \times W_{g-h}$ (called the cylinder set extension of h into g).

*Ballooning extension* Suppose h and g are subsets of $\mathfrak{X}$, $g \cap h = \emptyset$. For each $h \in W_h$, let $Bel_g(. \mid h)$ denote a belief function on $W_g$. Given these belief functions, we can construct the belief function on $W_{g \cup h}$ as follows (Smets 1991):

Let $Bel_{g \cup h}$ be the resulting belief function on $W_{g \cup h}$, called the *ballooning extension of* $Bel_g(. \mid h)$. Let $a \subseteq W_{g \cup h}$ and $g$ be the projection of $a \cap \{h\}^{\uparrow (g \cup h)}$ for g. Then

$$m_{g \cup h}(a) = \prod \{m_g(g \mid h) \mid h \in W_h\} \quad (2.1)$$

## 2.2 COMBINATION AND MARGINALIZATION

In order to define the combination and marginalization formally, we need to introduce a new representation for belief functions and utility valuations:

Suppose $\upsilon$ is a valuation for h, $\upsilon$ can always be represented by a set of pair $(a, \mu(a))$, where $a$ is a non-empty subset of $W_h$ and $\mu$ is mapping from $a$ to the set of real number. For each element x of $a$, we use $\mu(x, a)$ to represent its corresponding value. As this representation can be regarded as extension of bpa, then we call $a$ in this representation as an *extended focal element*.

If $\pi$ is a utility valuation for h defined as before, then $\pi$ can be represented the set of one pair $(W_h, \mu(W_h))$, where for each $x \in W_h$, $\mu(x, W_h) = \pi(x)$.

If bel is a belief function for h, then it can be represented as the set of $(a, \mu(a))$, where $a$ is a focal element of bel and for every x of $a$, $\mu(x, a) = m(a)$. We thus use $\mu(a)$ to represent any value of $\mu(x, a)$.

If p is a probability for h, then it can be represented as the set of $(a, \mu(a))$, where for each $(a, \mu(a))$, $a$ is a singleton set, and $\mu(x, a) = p(x)$.

*Combination* Suppose h and g are subsets of $\mathfrak{X}$, $g \cap h \neq \emptyset$. Let $\upsilon_i$, a valuation for g, be represented as a set of $(a, \mu(a))$, $\upsilon_j$, a valuation for h, be represented as a set of $(b, \mu(b))$, and the resulting valuation $\upsilon_i \otimes \upsilon_j$, a valuation for $g \cup h$, be represented as a set of $(c, \mu(c))$, where c is a non-empty subset of $W_{g \cup h}$ and $c \subseteq \{(a^{\uparrow (g \cup h)} \cap b^{\uparrow (g \cup h)})\}$. The definition of combination depends on the type of the valuations being combined.

If both of $\upsilon_i$ and $\upsilon_j$ are belief functions, then their combination is defined by Dempster's rule of combination. Formally, $\upsilon_i \otimes \upsilon_j$ is obtained by:

$$\mu(x, c) = K \sum \{(\mu_i(x^{\downarrow h}, a) \mu_j(x^{\downarrow g}, b)) \mid a^{\uparrow (g \cup h)} \cap b^{\uparrow (g \cup h)} = c\} \quad (2.2)$$

where

$$K = 1 - \sum \{\mu_i(a) \mu_j(b) \mid ((a^{\uparrow (g \cup h)} \cap b^{\uparrow (g \cup h)}) = \emptyset\}$$

If neither of $\upsilon_i$ and $\upsilon_j$ are belief functions, then $\upsilon_i \otimes \upsilon_j$ is obtained by:

$$\mu(x, c) = \sum \{(\mu_i(x^{\downarrow h}, a) + \mu_j(x^{\downarrow g}, b)) \mid a^{\uparrow (g \cup h)} \cap b^{\uparrow (g \cup h)} = c\} \quad (2.3)$$

Otherwise, $\upsilon_i \otimes \upsilon_j$ is obtained by:

$$\mu(x, c) = \sum \{(\mu_i(x^{\downarrow h}, a) \mu_j(x^{\downarrow g}, b)) \mid a^{\uparrow (g \cup h)} \cap b^{\uparrow (g \cup h)} = c\} \quad (2.4)$$

From the definition above, we have the following properties:
(1) If both $\upsilon_i$ and $\upsilon_j$ are belief functions, then $\upsilon_i \otimes \upsilon_j$ is also a belief function.
(2) If both of $\upsilon_i$ and $\upsilon_j$ are utility valuations, then $\upsilon_i \otimes \upsilon_j$ is also a utility valuation;
(3) If one of $\upsilon_i$ and $\upsilon_j$ is not a belief function, then $\upsilon_i \otimes \upsilon_j$ is not a belief function. The resulting valuation could be a utility valuation or a collection of subset-utility pairs.
(4) The combination is commutative;
(5) The combination for belief functions is associative, so is that for utility valuations.
(6) The combination for a mixture of belief functions and utility valuations is not associative.

In case (6), we defined combination such that the utility valuations are combined before the belief functions. Formally, suppose $\pi_1, ..., \pi_m$ are utility valuations and $bel_1, ..., bel_n$ are belief functions. Then $(\otimes \{\pi_1, ..., \pi_m, bel_1, ..., bel_n\})$ denotes $(\otimes \{\pi_1, ..., \pi_m\} \otimes (\otimes \{bel_1, ..., bel_n\})$.

For the definition of marginalization, we need to use a weighting factor $\lambda$, which is real number in [0, 1]. The role of $\lambda$ in marginalization is similar to the probabilistic interpretation of an assumption that disambiguates decision problems represented with belief functions (Strat 1990).

*Marginalization* Suppose h is a subset of $\mathfrak{X}$ containing variable X, and $\upsilon$ is a valuation for h, represented as a set of $(a, \mu(a))$. The definition of marginalization depends on the type of variables being eliminated instead of on the type of the valuation being marginalized. Let the *marginal of $v$ for $h-\{X\}$*, denoted by $v^{\downarrow(h-\{X\})}$ is a valuation for $h-\{X\}$, be represented by a set of $(c, \mu(c))$, it is defined by follows:

If X is a decision variable, then for any $x \in c$,
$$\mu(x, c) = \sum \{MAX[\mu(y, a) \mid y \in a, y^{\downarrow(h-\{X\})} = x]$$
$$\mid a^{\downarrow(h-\{X\})} = c\} \quad (2.5)$$

If X is a random variable, then for any $x \in c$,
$$\mu(x, c) = \sum \{\lambda MAX[\mu(y, a) \mid y \in a, y^{\downarrow(h-\{X\})} = x]$$
$$+(1-\lambda)MIN[\mu(y, a) \mid y \in a, y^{\downarrow(h-\{X\})} = x]$$
$$\mid a^{\downarrow(h-\{X\})} = c\} \quad (2.6)$$

From the definition above, we have the following properties:

(1) Suppose h is a subset of $\mathcal{X}$ containing decision variable $D_1$ and $D_2$, and $v$ is a non-belief function valuation for h. Then
$$(v^{\downarrow(h-\{D_1\})})^{\downarrow(h-\{D_1,D_2\})} =$$
$$(v^{\downarrow(h-\{D_2\})})^{\downarrow(h-\{D_1,D_2\})}$$

(2) Suppose h is a subset of $\mathcal{X}$ containing random variable $R_1$ and $R_2$, and $v$ is a non-belief function valuation for h. Then
$$(v^{\downarrow(h-\{R_1\})})^{\downarrow(h-\{R_1,R_2\})} \neq$$
$$(v^{\downarrow(h-\{R_2\})})^{\downarrow(h-\{R_1,R_2\})} \quad (2.7)$$

But if $v$ is a belief function, then
$$(v^{\downarrow(h-\{R_1\})})^{\downarrow(h-\{R_1,R_2\})} =$$
$$(v^{\downarrow(h-\{R_2\})})^{\downarrow(h-\{R_1,R_2\})}$$

(3) Suppose $h \subseteq \mathcal{X}$, $R \in h$, $R \in \mathcal{X}_R$, and $\rho$ is a bpa for h. if $\rho$ is the ballooning extension of $Bel_R(.\mid g)$ where $g \in \mathcal{W}_{h-\{R\}}$, then $\rho^{\downarrow(h-\{R\})}$ is a vacuous belief function for $h-\{R\}$. Thus, $\rho$ is called a *conditional potential(belief function) for R given $h-\{R\}$*

The deletion sequence respect to the precedence constraint. The inequality of (2.7) is because there are two random variables $R_1$ and $R_2$ in the deletion sequence such that $X_1 > R_1 > X_2$ and $X_1 > R_2 > X_2$. In order to avoid the inequality of (2.7), the precedence relation need an additional constraint:

(p5): For any two random variables $R_1$ and $R_2$, there exists at least a decision variable D such that $R_1 > D > R_2$ or $R_2 > D > R_1$ of there exist decision variables in the valuation network.

(p5) is consistent to the role of $\lambda$. Similar to the probabilistic interpretation for making assumption in (Strat 1990), $\lambda$ can be used only before the action must be chosen for a decision variable, i.e., only one random variable is allowed to be deleted before each decision variable as $\lambda$ is used once each time when a random variable is deleted.

If there are no decision variables and utility valuations in VBS, combination is then reduced to Dempster's rule of combination, and marginalization (2.6) is then reduced to the following:
$$\mu(c) = \sum \{\mu(a) \mid a^{\downarrow(h-\{X\})} = c\} \quad (2.8)$$

Thus, the VBS is reduced to an evidential system for propagating belief functions.

### 2.3 SOLVING VBS

Solving Bayesian decision problems in VBS is based on the criterion of maximizing expected payoff. The presented calculus is essentially based on the generalization for expectation operation for belief functions proposed by Strat (1990). Let us look at the solution for the canonical decision problem. As shown in Fig. 1, $\Delta_c = \{\{D\}, \{R\}, \{\mathcal{W}_D, \mathcal{W}_R\}, \{\pi\}, \{\rho\}, \rightarrow\}$ where $\rho$ is a conditional potential(belief function) for R given $\{D\}$. Let $\pi \otimes \rho$ be represented as $(a, \mu(a))$. Based on the generalization for expectation operation for belief functions, the expected utility interval is computed by:
$$[\sum(MIN(\mu((d,r),a) \mid r \in \mathcal{W}_R)),$$
$$\sum(MAX(\mu((d,r),a) \mid r \in \mathcal{W}_R))] \quad (2.9)$$

But an interval of expected values is not very satisfactory when we have to make a decision. Thus, additional assumptions need to be made to compute a unique expected utility. To this end, a weighting factor $\lambda$ to resolve the ambiguity is needed, and expected value (associated with an optimal act $d^*$ with respect to $\lambda$) is $((\pi \otimes \rho)^{\downarrow\{D\}})^{\downarrow\varnothing}(\blacklozenge)$, act $d^*$ is optimal if $(\pi \otimes \rho)^{\downarrow\{D\}}(d^*) = ((\pi \otimes \rho)^{\downarrow\{D\}})^{\downarrow\varnothing}(\blacklozenge)$. Here, $(\pi \otimes \rho)^{\downarrow\{D\}}$ is a utility valuation.

If $\rho$ in $\Delta_c$ is defined as a conditional potential for probability (Shenoy 1991a), the computation of expected utility interval is reduced to the computation of unique expected utility, and the operators of combination and marginalization are reduced to those for Bayesian probability theory.

Formally, we have the following theorem for solving VBS in the case belief functions:

**Theorem 2**: (Xu 1992) In the case of belief functions, suppose $\Delta = \{\mathcal{X}_D, \mathcal{X}_R, \{\mathcal{W}_X\}_{X \in \mathcal{X}}, \{\pi_1, ..., \pi_m\}, \{\rho_1, ..., \rho_n\}, \rightarrow\}$ is a well-defined VBS representation (including (p5) for precedence relation), and satisfies the assumption 1. Using the decision calculus defined above, we can also use fusion algorithm to solve such VBS. See (Xu 1992) for proofs.

### 2.4 EXAMPLE:

The oil wildcatter's problem (Strat 1990)

An oil wildcatter must decide either to drill (d) of not to drill (~d). He is uncertain whether the hole is dry (dr), wet (we) or soaking (so). Drilling a hole costs $70,000. The payoffs for hitting a soaking, a wet or a dry hole are $270,000, $120,000, and $0, respectively. At a cost of $10,000, the wildcatter can make an electronic test that is related to the well capacity of the oil. The result is shown below:

| Prob. | Test Result | Capacity |
|---|---|---|
| 0.5 | Red (re) | Dry |
| 0.2 | Yellow(ye) | Dry or Wet |
| 0.3 | Green(gr) | Wet or Soaking |

The VBS representation of the problem according to the information above is illustrated in Fig. 2. The valuation network consists of two decision variables D with frame {d, ~d} and T with frame {t, ~t}, two random variables R (test results) with frame {re, ye, gr, nr}, where nr(no result) represents the state when no test is taken, and O (the state of the oil) with frame {dr, we, so}. We also have the precedence relation: T→R, R→D, D→O.

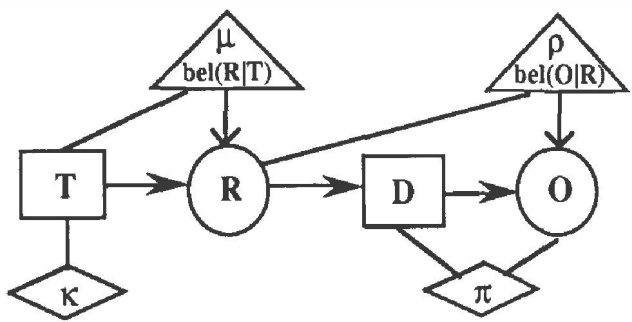

Fig 2: Graphical VBS representation for the oil wildcatter's problem

There are two utility valuations and conditional potentials in the network. The utility valuations are as follows:

| $a \subseteq \Psi_{\{D,O\}}$ | | $\pi$ |
|---|---|---|
| d | dr | -70,000 |
| d | we | 50,000 |
| d | so | 200,000 |
| ~d | dr | 0 |
| ~d | we | 0 |
| ~d | so | 0 |

| $a \subseteq \Psi_{\{T\}}$ | $\kappa$ |
|---|---|
| t | -10,000 |
| ~t | 0 |

The two potentials are defined by the conditional belief functions:
Bel(Result I Test):
   m({re} | t) = 0.5, m({ye} | t) = 0.2, m({gr} | t) = 0.3;
   m({nr} | ~t) = 1;
Bel (Oil I Test):
   m({dr} | re) = 1;
   m({dr, we} | ye) = 1;
   m({we, so} | gr) = 1;
   m({dr} | nr) = 0.5, m({dr, we} | nr) = 0.2,
   m({we, so} | nr) = 0.3

Their corresponding ballooning extensions are:

| $a \subseteq \Psi_{\{T,R\}}$ | $\mu$ |
|---|---|
| t   re<br>~t  nr | 0.5 |
| t   ye<br>~t  nr | 0.2 |
| t   gr<br>~t  nr | 0.3 |

| $b \subseteq \Psi_{\{R,O\}}$ | $\rho$ |
|---|---|
| re  dr<br>ye  dr<br>ye  we<br>gr  we<br>gr  so<br>nr  dr | 0.5 |
| re  dr<br>ye  dr<br>ye  we<br>gr  we<br>gr  so<br>nr  dr<br>nr  we | 0.2 |
| re  dr<br>ye  dr<br>ye  we<br>gr  we<br>gr  so<br>nr  we<br>nr  so | 0.3 |

It is easy to prove that the representation is well-defined. Suppose the weighting factor $\lambda=0.5$. Then, according the solution method, we can compute the expected value by $(((((\kappa \otimes \pi) \otimes (\rho \otimes \mu))^{\downarrow\{T,R,D\}})^{\downarrow\{T,R\}})^{\downarrow\{T\}})^{\downarrow \emptyset}$. By using the fusion algorithm, the expected utility is computed by: $(((((\rho \otimes \pi)^{\downarrow\{R,D\}})^{\downarrow\{R\}}) \otimes \mu)^{\downarrow\{T\}} \otimes \kappa)^{\downarrow \emptyset}$ The computation is illustrated in the tables of the Appendix. With $\lambda=0.5$, the expected value is $27.500. An optimal strategy can be constructed from the information in $\Psi_T$ and $\Psi_D$ (Shenoy, 1991a,d). From $\Psi_T$, it can be seen that the wildcatter should do the test. And from $\Psi_D$, it can be seen that if the test result is red, then optimal decision is not to drill; If the result is yellow or green, then optimal decision is to drill. The expected payoff value and strategy change according to the value of $\lambda$.

358 Xu

## 3. CONCLUSIONS

We have presented and discussed a decision calculus for D-S theory in the extended framework of VBS. The presented calculus can be regarded as a generalization of the calculus for the Bayesian probability theory. If there are no random variables in the problem, which is thus an optimization problem, the solution technique reduces to dynamic programming (Shenoy, 1991c,d); If there are no decision variables, then our objective may become to find the marginal of some variables and the solution technique reduces to the one for belief function propagation in an evidential system. The presented calculus has some limitations because only the conditional potentials are allowed in the representation. For the case of probability theory, VBS representation can directly represent arbitrary probability, and the division operation is introduced to the solution method for such a case. Extension to this case for D-S theory remains a topic for future research.

We have also shown that the presented calculus needs a weighting factor for marginalization. This idea is similar to Strat's method for making the additional assumption when sufficient information is not available for making the decision. Although this calculus may not be the best one for decision making, as Strat commented on his method, our interest in this paper is to develop a decision calculus for D-S theory which can benefit the advantages of the VBS representation and solution method for the decision problems, especially of the fusion algorithm which can reduce the complexity of computation.

### Acknowledgements

I am indebted to Professor Philippe Smets for his encouragement and support. I would like to thank the belief group at IRIDIA which is Yen-Teh Hsia, Robert Kennes, Victor Poznanski, Alessandro Saffiotti. Especially, I am very grateful to Yen-Teh Hsia for his careful reading and valuable comments on the paper. This research has been supported by a grant of IRIDIA, Université libre de Bruxelles.

**APPENDIX :**
   (tables of the computations for the example)

Table 1: computation of $(\rho \otimes \pi)^{\downarrow \{R,D\}}$.

Let $c = a^{\uparrow\{R,D,O\}} \cap b^{\uparrow\{R,D,O\}}$, where $a, b$ are the extended focal elements of $\pi$ and $\rho$, respectively. $\lambda$ is used in marginalization.

| $c \subseteq W_{\{R,D,O\}}$ | $\rho$ | $\pi$ | $\rho \otimes \pi$ | $(\rho \otimes \pi)^{\downarrow\{R,D\}}$ |
|---|---|---|---|---|
| re  d  dr  | 0.5 | -70,000 | -35,000 | -35,000 |
| re ~d  dr  |     | 0       | 0       | 0       |
| ye  d  dr  |     | -70,000 | -35,000 | -5,000  |
| ye  d  we  |     | 50,000  | 25,000  |         |
| ye ~d  dr  |     | 0       | 0       | 0       |
| ye ~d  we  |     | 0       | 0       |         |
| gr  d  we  |     | 50,000  | 25,000  | 62,500  |
| gr  d  so  |     | 200,000 | 100,000 |         |
| gr ~d  we  |     | 0       | 0       | 0       |
| gr ~d  so  |     | 0       | 0       |         |
| nr  d  dr  |     | -70,000 | -35,000 | -35,000 |
| nr ~d  dr  |     | 0       | 0       | 0       |
| re  d  dr  | 0.2 | -70,000 | -14,000 | -14,000 |
| re ~d  dr  |     | 0       | 0       | 0       |
| ye  d  dr  |     | -70,000 | -14,000 | -2,000  |
| ye  d  we  |     | 50,000  | 10,000  |         |
| ye ~d  dr  |     | 0       | 0       | 0       |
| ye ~d  we  |     | 0       | 0       |         |
| gr  d  we  |     | 50,000  | 10,000  | 25,000  |
| gr  d  so  |     | 200,000 | 40,000  |         |
| gr ~d  we  |     | 0       | 0       | 0       |
| gr ~d  so  |     | 0       | 0       |         |
| nr  d  dr  |     | -70,000 | -14,000 | -2,000  |
| nr  d  we  |     | 50,000  | 10,000  |         |
| nr ~d  dr  |     | 0       | 0       | 0       |
| nr ~d  we  |     | 0       | 0       |         |
| re  d  dr  | 0.3 | -70,000 | -21,000 | -21,000 |
| re ~d  dr  |     | 0       | 0       | 0       |
| ye  d  dr  |     | -70,000 | -21,000 | -3,000  |
| ye  d  we  |     | 50,000  | 15,000  |         |
| ye ~d  dr  |     | 0       | 0       | 0       |
| ye ~d  we  |     | 0       | 0       |         |
| gr  d  we  |     | 50,000  | 15,000  | 37,500  |
| gr  d  so  |     | 200,000 | 60,000  |         |
| gr ~d  we  |     | 0       | 0       | 0       |
| gr ~d  so  |     | 0       | 0       |         |
| nr  d  we  |     | 50,000  | 15,000  | 37,500  |
| nr  d  so  |     | 200,000 | 60,000  |         |
| nr ~d  we  |     | 0       | 0       | 0       |
| nr ~d  so  |     | 0       |         |         |

Table 2: computation of $\tau^{\downarrow\{R\}}$ where $\tau = (\rho \otimes \pi)^{\downarrow\{R,D\}}$. $\Psi_D$ is stored when D is deleted.

| $c \subseteq W_{\{R,D\}}$ | $\tau^{\downarrow\{R,D\}}$ | $\tau^{\downarrow\{R\}}$ | $\Psi_D$ |
|---|---|---|---|
| re  d  | -70,000  |         |    |
| re ~d  | 0        | 0       | ~d |
| ye  d  | -10,000  |         |    |
| ye ~d  | 0        | 0       | ~d |
| gr  d  | 125,000  | 125,000 | d  |
| gr ~d  | 0        |         |    |
| nr  d  | 500      | 500     | d  |
| r  ~d  | 0        |         |    |

Table 3: computation of $(\tau^{\downarrow\{R\}} \otimes \mu)^{\downarrow\{t\}}$ where $\tau = (\rho \otimes \pi)^{\downarrow\{R,D\}}$.

Let $c = a^{\uparrow\{R,T\}} \cap b^{\uparrow\{R,T\}}$, where $a, b$ are the extended focal elements of $\mu$ and $\tau^{\downarrow\{R\}}$, respectively. $\lambda$ is not needed in marginalization because $\mu$ is a probability.

| $c \subseteq W_{\{R,T\}}$ | $\mu$ | $\tau^{\downarrow\{R\}}$ | $\tau^{\downarrow\{R\}} \otimes \mu$ | $(\tau^{\downarrow\{R\}} \otimes \mu)^{\downarrow\{t\}}$ |
|---|---|---|---|---|
| re   t | 0.5 | 0       | 0      | 0      |
| nr  ~t |     | 500     | 250    | 250    |
| ye   t | 0.2 | 0       | 0      | 0      |
| nr  ~t |     | 500     | 100    | 100    |
| gr   t | 0.3 | 125,000 | 37,500 | 37,500 |
| nr  ~t |     | 500     | 150    | 150    |

Table 4: computation of $(\upsilon \otimes \kappa)^{\downarrow\emptyset}$ where $\upsilon = (\tau^{\downarrow\{R\}} \otimes \mu)^{\downarrow\{t\}}$. $\Psi_T$ is stored when T is deleted.

| $a \subseteq W_{\{T\}}$ | $\upsilon$ | $\kappa$ | $\upsilon \otimes \kappa$ | $\tau^{\downarrow\emptyset}(\bullet)$ | $\Psi_T$ |
|---|---|---|---|---|---|
| t  | 37,500 | -10,000 | 27,500 | 27,500 | t |
| ~t | 500    | 0       | 500    |        |   |